\documentclass[10pt,twocolumn,letterpaper]{article}

\usepackage[pagenumbers]{cvpr} 

\usepackage{graphicx}
\usepackage{amsmath}
\usepackage{amssymb}
\usepackage{booktabs}

\usepackage{times}
\usepackage{epsfig}
\usepackage{amsfonts}
\usepackage{bm}
\usepackage{float}
\usepackage{newtxmath}
\usepackage{pifont}
\usepackage{multirow}

\usepackage[pagebackref,breaklinks,colorlinks]{hyperref}

\usepackage[capitalize]{cleveref}
\crefname{section}{Sec.}{Secs.}
\Crefname{section}{Section}{Sections}
\Crefname{table}{Table}{Tables}
\crefname{table}{Tab.}{Tabs.}

\newcommand{\cmark}{\ding{51}}%
\newcommand{\xmark}{\ding{55}}%

\def\cvprPaperID{*****} 
\def\confName{CVPR}
\def\confYear{2023}

\makeatletter
\def\@fnsymbol#1{\ensuremath{\ifcase#1\or \dagger\or \ddagger\or
   \mathsection\or \mathparagraph\or \|\or **\or \dagger\dagger
   \or \ddagger\ddagger \else\@ctrerr\fi}}
    \makeatother

\begin{document}

\title{Learning to Summarize Videos by Contrasting Clips}

\author{Ivan Sosnovik\thanks{Work was done prior to joining Amazon}\\
UvA-Bosch Delta Lab\\
University of Amsterdam\\
{\tt\small i.sosnovik@uva.nl}
\and
Artem Moskalev\\
UvA-Bosch Delta Lab\\
University of Amsterdam\\
{\tt\small a.moskalev@uva.nl}
\and
Cees Kaandorp\\
Institute of Informatics\\
University of Amsterdam\\
{\tt\small cees.kaandorp@gmail.com}
\and
Arnold Smeuldes\\
UvA-Bosch Delta Lab\\
University of Amsterdam\\
{\tt\small a.w.m.smeulders@uva.nl}
}
\maketitle

\begin{abstract}
    Video summarization aims at choosing parts of a video that narrate a story as close as possible to the original one. Most of the existing video summarization approaches focus on hand-crafted labels. As the number of videos grows exponentially, there emerges an increasing need for methods that can learn meaningful summarizations without labeled annotations. In this paper, we aim to maximally exploit unsupervised video summarization while concentrating the supervision to a few, personalized labels as an add-on. To do so, we formulate the key requirements for the informative video summarization. Then, we propose contrastive learning as the answer to both questions. To further boost Contrastive video Summarization (CSUM), we propose to contrast top-k features instead of a mean video feature as employed by the existing method, which we implement with a differentiable top-k feature selector. Our experiments on several benchmarks demonstrate, that our approach allows for meaningful and diverse summaries when no labeled data is provided.
\end{abstract}
\section{Introduction}
\label{sec:csum_introduction}

\begin{figure}[t!]
    \centering
      \includegraphics[width=0.9\linewidth]{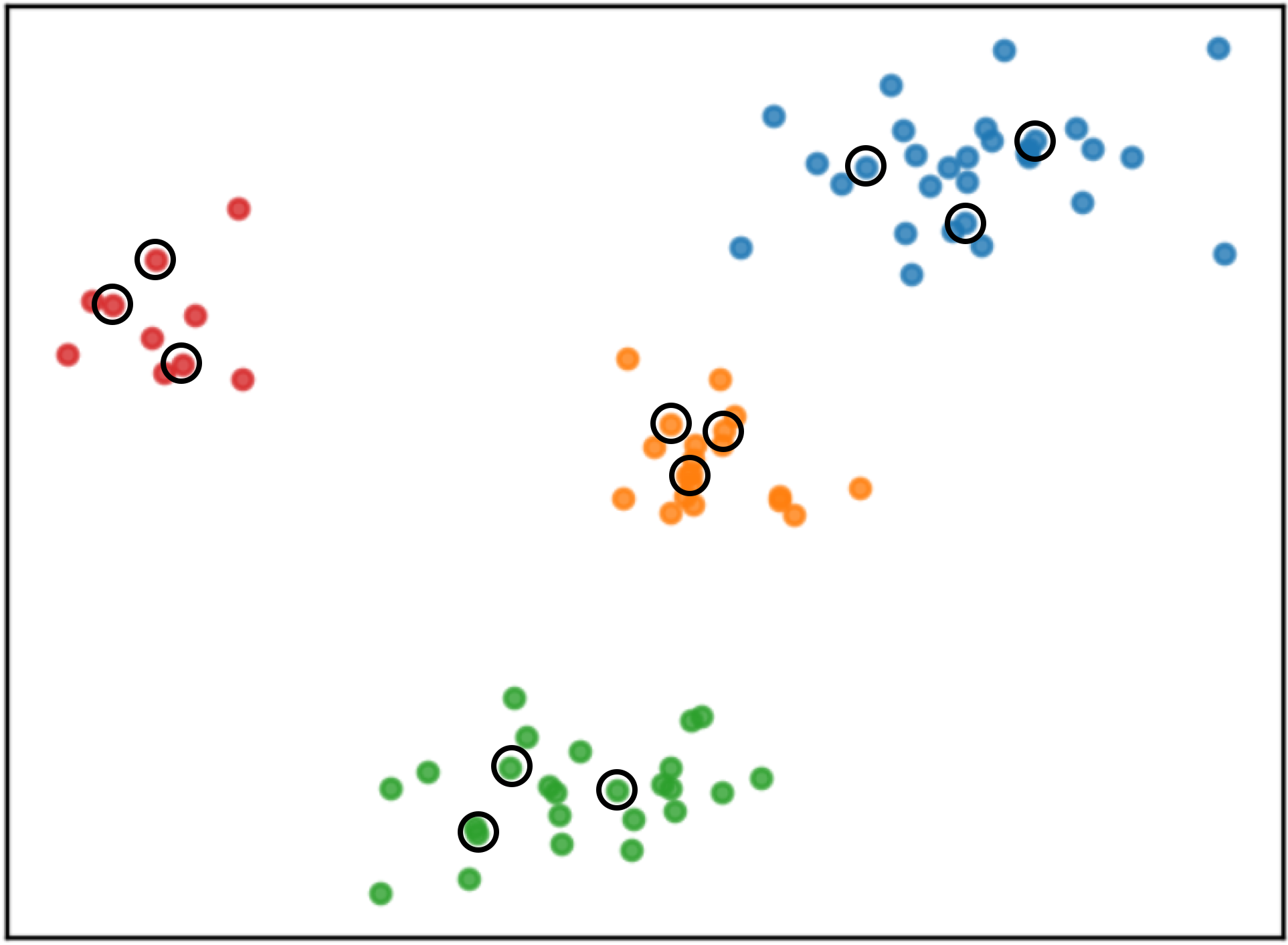}
    \caption{The t-SNE of the feature space of example clips and the learned summaries (black circles) for these clips. Finding a good summary in the feature space does not boil down to the centroids of corresponding feature distributions, but rather consists of finding samples that informatively describe the whole input sequence.}
    \label{fig:csum_tsne}
\end{figure}

Video summarization aims at choosing parts of a video that narrate a story as close as possible to the original one. In this day and age, video streaming without personalized recommendations is almost gone. Current recommendations select a fixed preview provided by the distributor on the basis of past preferences. We aim to go one step further and to provide personalized previews. Apart from better video selection for streaming, it also opens possibilities for better video editing, ad creation, and edge-device software development. 

Existing approaches for video summarization focus on supervised summarization \cite{gong2014diverse,li2018local,rochan2018video,zhang2016video,zhou2018deep,zhao2017hierarchical}. With a growing number of videos, supervised summarization may still be somewhat affordable for the distributor. When the number of videos grows exponentially, the supervised model is not sustainable. And, from the standpoint of the user labeling can be applied only in very moderate amounts. In this paper, we aim to maximally exploit unsupervised video summarization while concentrating the supervision on a few personalized labels as an add-on.
\begin{figure*}[h!]
    \centering
      \includegraphics[width=0.9\linewidth]{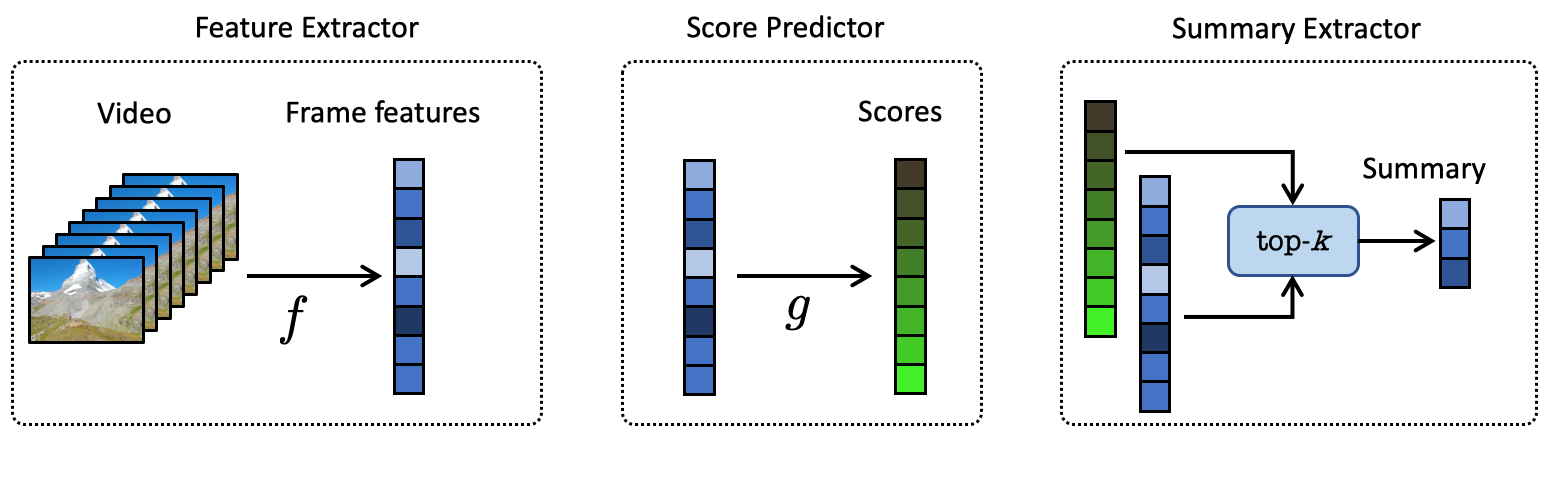}
    \caption{The main building blocks of our framework for video summarization. Feature extractor $f$ transforms an arbitrary-length sequence of frames into a sequence of features. They are then used to predict a set of scores by using a function $g$. The summary extractor block uses both sequences to extract a set of $k$ frames with the highest scores.}
    \label{fig:csum_building_blocks}
\end{figure*}

Unsupervised video summarization techniques were developed in the pre-deep-learning era, when no large labeled datasets were available \cite{gong2000video}. They are still being used these days as labeling the full spectrum of possible videos is no longer possible \cite{mademlis2018regularized}. Regardless of the method of video analysis, with deep learning or not, the main reasoning for selecting a good summary is left unchanged. The summary must be a compressed representation of the original video while being closer in content to the source than to other videos \cite{mundur2006keyframe,gharbi2019key}. Two core questions remain: how to summarize a video while preserving most information in the video, and how to measure distances between two videos on the basis of their content? Formulating the video summarization in this way, in this paper, we propose contrastive learning \cite{chen2020simple} as the answer to both questions. Contrastive learning was designed to handle compression and metric learning in one go. We propose that neural networks can be trained to optimize a contrastive loss which represents the core requirements of a good summary.

Contrastive learning has been successfully applied in image and video classification \cite{chen2020simple,roy2021spatiotemporal,dave2022tclr}. In classification, the contrastive loss is evaluated by comparing descriptive feature vectors of equal size \cite{jing2020self}. In video summarization, the comparison is between a vector describing the full video and a vector describing the summary. As a consequence, the vectors will not be equal in size, and hence cannot be compared directly by common contrastive losses. To overcome the inequality in size, the common approach for comparing vector representations of videos and their summaries is to use their time-averaged representations \cite{badamdorj2022contrastive}. In this work, we note that such an approach is invariant to a wide range of transformations and does not account for important moments of high information in the video. When taking an average, the summary is adequate when the video develops slowly like a game of snooker or a sit-com interview but expected to be less adequate when there are short moments of great significance. While various architectural solutions were proposed to improve over quality by average \cite{rochan2018video}, we propose that a combination of a well-chosen loss function and training approach suffices to avoid the unwanted invariances for a wide range of backbones.

In this paper, we focus on developing a simple, flexible, yet efficient recipe for contrastive training of deep-learning-based video summarizers. We start from the principle of maximum information preservation. We demonstrate that the ensuing loss function and maximization process preserves important information while avoiding undesired invariances. 
Our main contributions are the following:
\begin{itemize}
    \item From the requirements of video summaries we propose a method for contrastive learning of video summaries with no need for labeling.
    \item We propose implementations of the main building blocks which are required to convert any video-analysis network into a summarizer.
    \item We demonstrate the advantage of contrastive video summarizers on popular benchmarks for a set of backbone architectures over their original training methods. We also demonstrate how it can be used for video highlight selection with a slight modification.
\end{itemize}
\section{Related Work}
\label{sec:csum_related}

\begin{figure*}[t!]
    \centering
        \includegraphics[width=0.98\linewidth]{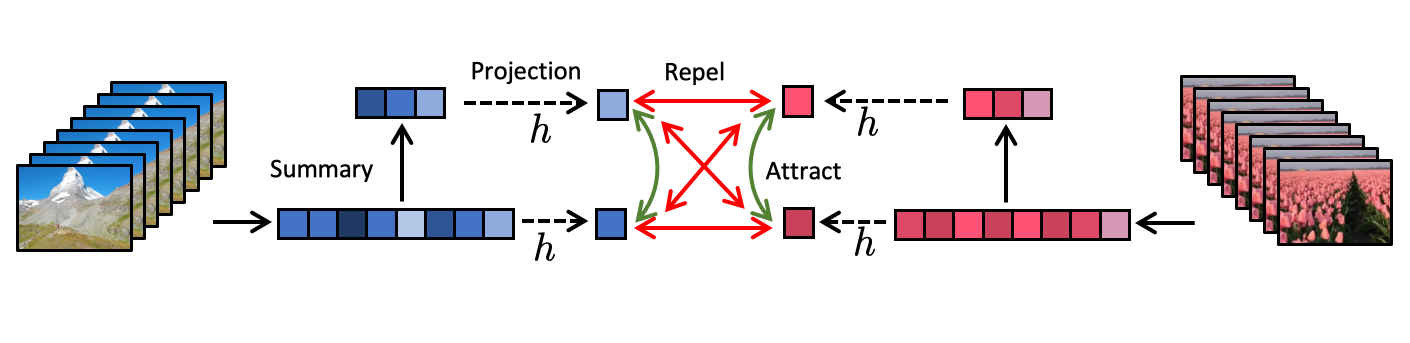}
    \caption{An illustration of the proposed contrastive training pipeline. Given two videos, the features are first calculated for them and then a summary is extracted. Both summaries and original videos are projected by a neural network $h$ to a hidden space afterward. The whole pipeline is trained to attract the projections of summaries to the projections of the original videos and to repel them from other summaries and videos.}
    \label{fig:csum_pipeline}
\end{figure*}

\paragraph{Supervised Methods}
With the rise of deep learning, a wide range of papers has considered video summarization as a regression problem. 
In such a paradigm a neural network is used to take frames of the video and predict their importance scores so that the top-scored frames form the summary.
In \cite{zhang2016video} the authors use Recurrent Neural Networks (RNNs) to combine the temporal information from the video with the content of each frame to successfully predict frames' scores.
Alternatively, in \cite{rochan2018video} and \cite{fajtl2018summarizing} convolutional and attention-based architecture was proposed to improve the quality of predictions.
To effectively combine information about videos from multiple scales, hierarchical models were proposed \cite{zhao2017hierarchical,zhao2018hsa}. By using hierarchical RNNs, 
models benefit from considering the video as a whole, as a set of short clips and as a sequence of individual frames at the same time. It allows to create summaries of less-contract granularity than before.
While these methods demonstrated the great success of deep neural networks for video summarization, manually labeled annotations are required for their training. 
It makes it impossible the scaling of such methods to long videos, movies and streams of videos that are constantly being uploaded on the major video services. For these reasons, we focus on methods that do not rely on human-annotated labels. 

\paragraph{Unsupervised Methods}
Early-day methods for video summarization relied on heuristics designed by a human. 
The heuristics were designed to satisfy the main requirements for video summaries such as representativeness and diversity, justified in \cite{ngo2003automatic,de2011vsumm}.
In \cite{mundur2006keyframe,de2011vsumm,kuanar2013video} the authors clustered frames and use the centroids to form a summary. 
The authors of \cite{cong2011towards,mei20142} formulate video summarization as a sparse dictionary selection problem. 
Later, in the deep-learning era, video summarization was approached from the perspective of adversarial training \cite{he2019unsupervised,mahasseni2017unsupervised} or in the reinforcement learning paradigm \cite{zhou2018deep}. We draw inspiration from the pre-deep-learning era methods. By starting from the reasoning of video summarization, we demonstrate that we can satisfy the main requirements by formulating it as a contrastive learning problem that we can easily solve. 

\paragraph{Contrastive Learning}
Contrastive learning is an approach for performing self-supervised pretraining of a model by using a pre-text task. The model learns to attract representations that are meant to be close, and are thus called positive, and repel them from negative representations which are meant to be distant enough to distinguish between different objects \cite{hadsell2006dimensionality,tschannen2019mutual,chen2020simple,moskalev2022contrasting}. Various methods have been proposed for learning image-level \cite{caron2021emerging,chen2021exploring} and spatio-temporal models \cite{fernando2017self,goroshin2015unsupervised,badamdorj2022contrastive}. The current application of contrastive learning methods for video summarization is rather limited due to special architectural solutions dictated by the domain. In this work, we demonstrate an approach for contrastive learning for video summarization that does not rely on any specific backbones and allows one to use any model and framework of their choice.

\paragraph{Video Highlight Selection} 
Another popular approach for creating a compressed visual representation of videos is video highlight selection. While the summary has a fixed length, the highlights are not bounded in length but have a lower bound for the importance scores. Various methods have been proposed for solving this problem both from the supervised perspective \cite{sun2014ranking,xiong2019less,xu2021cross,hong2020mini,ye2021temporal,badamdorj2021joint,liu2022umt}, as well as in the unsupervised manner \cite{badamdorj2022contrastive}. In this paper, we demonstrate that with a slight modification of our video summarization framework, we can outperform modern video highlight selection models without significant transformations of the original pipeline.

\section{Method}
\label{sec:csum_method}

\subsection{Summary Requirements}
Video summarization is a very subjective task, as a manually labeled summary is biased towards the personal preferences of the annotator, assessor \cite{song2015tvsum}. However, it is possible to select several properties of a good summary that we would consider as summary requirements. They also give us hint on how to build an efficient model for video summarization.
\paragraph{Representativeness}
The composed summary should deliver the same message as the original video. As the summary is a compressed representation of the source, the loss of the original information is inevitable. However, we require a good summary to contain all the information necessary to distinguish between the original video and all other videos \cite{mahasseni2017unsupervised}. With no loss of generality, we can assume that each video contains a finite set of sub-videos each of which tells a separate narrative. Thus, the desired summary is a combination of sub-videos that is as close as possible to all of them at the same time, as well as distant enough from all sub-videos of other original videos. We suggest to learn summaries by selecting a set of sub-videos which we call clips which once they are projected to some hidden space minimize a variant of the triplet loss. As we want to develop a model for unsupervised summarization, it leads us to the framework of contrastive learning \cite{chen2020simple}.

\paragraph{Sparsity}
The original videos may come from various sources: be it video news, a video blog, several-hours-long online streams or a TV show. For all cases, the desired summary would be just several seconds long as it is the average amount of time a user can spend before deciding whether to watch it or to skip it. It leads us to the requirement of the sparsity of the resulting summary. While for short videos the desired summary may be around 15\% of the length \cite{gygli2014creating}, this ratio may drop significantly for longer videos. Thus, our model should be capable of choosing the very top segments of the video with a significant distinction from the rest. The problem of ranking items and selecting the top of them cannot be overestimated as it has significant limitations in the realm of deep learning especially when it comes to a very sparse output \cite{xie2020differentiable,petersen2022differentiable,goyal2018continuous}. In our approach, we should not directly rely on any heuristics for performing such an operation and should seek an as accurate as possible algorithmic implementation of it.

\paragraph{Diversity} 
Another important property of a video summary is the diversity among its frames. We may assume that for some videos and for some datasets it is possible to create a summary that will contain a lot of very similar frames and clips. Although it is possible for a user to understand what the video is about just by taking a look at one frame, it is still desired to have a summary with a higher diversity of visual information. If we consider two models which satisfy the above-mentioned requirements, we want to select the model which selects diverse summaries over uniform summaries. It shows us that the function which we will use to measure the distance between videos should not be invariant to the spread of the frames. In other words, it should take into account not only a single frame and the general content of videos but also the variations inside them.

\begin{figure}[t!]
    \centering
      \includegraphics[width=\linewidth]{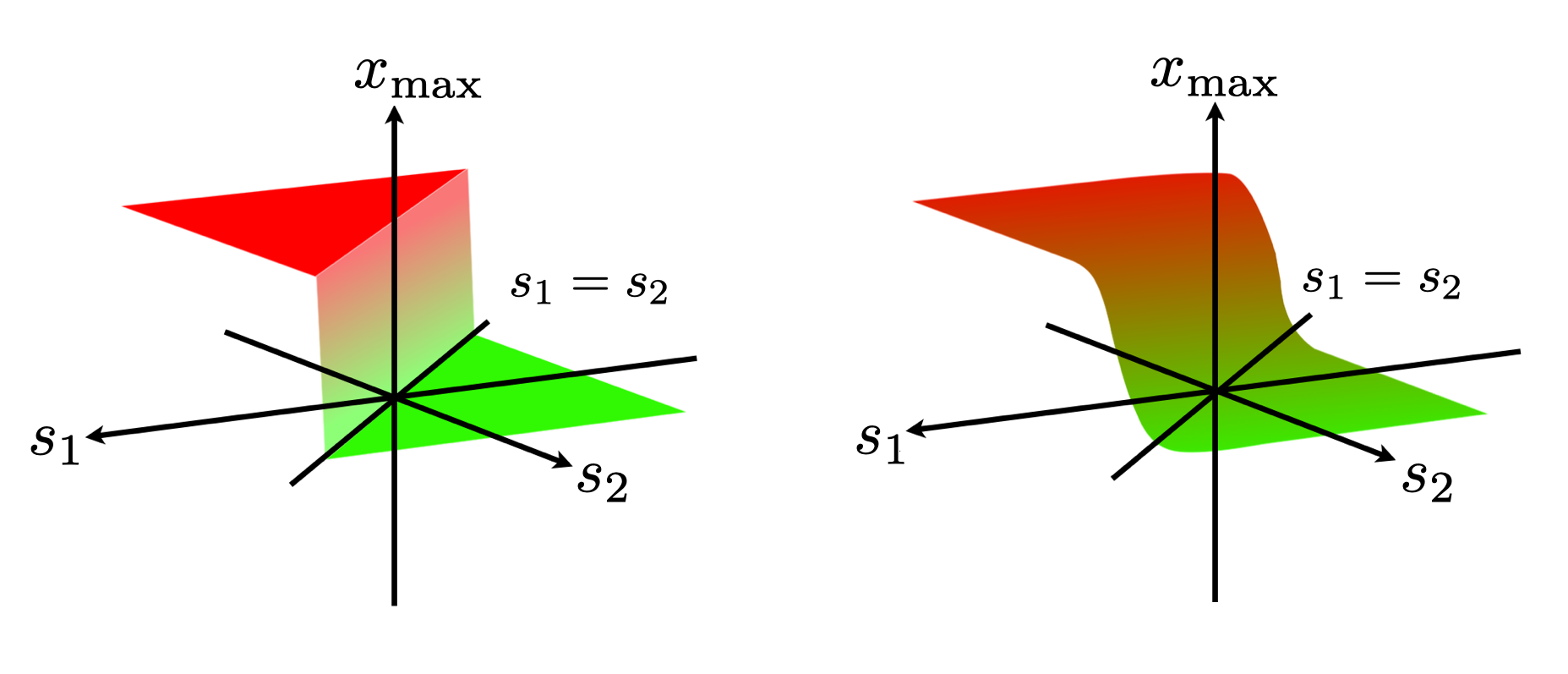}
    \caption{Illustration of frame selection based on the scores $s_1$ and $s_2$. Left: the original step function $x_\text{max} = x_1$ if $s_1 > s_2$ and $x_\text{max}=x_2$ otherwise. Right: a relaxed version with a smooth replacement for the step function.}
    \label{fig:csum_differentiation}
\end{figure}

\subsection{Contrastive Summarization}
A wide range of trainable video summarizers can be decomposed into the following three blocks: features extractor $f$, score predictor $g$ and summary extractor (see Figure \ref{fig:csum_building_blocks}). From the summary requirement, we generated several requirements for the video summarization pipeline. And none of them are related to the feature extractor or the score predictor. Thus, we assume that these two blocks are the free parameters of our framework. Once a feature extractor and a score predictor are chosen, we train their parameters by performing a variant of contrastive learning.

During training, we consider two videos (see Figure \ref{fig:csum_pipeline}). Each of the videos is processed with the feature extractor and the score predictor functions. After that, for each of the videos a summary is generated. We choose the parameters of the networks $f$ and $g$ by training them to generate video and summary embeddings s.t. the summary attracted to its source video is repelled from any other videos and summaries. It is done by minimizing the following loss \cite{chen2020simple}:
\begin{equation}
    \label{eq:csum_contrastive_loss}
    \mathcal{L} = \sum_{z, z_{+}}-\log \frac{\exp(\text{dist}(z, z_{+})/\tau)}{\sum_{z_{-}}\exp(\text{dist}(z, z_{-})/\tau)}
\end{equation}
where $z, z_{-}, z_{+}$ are embeddings for the anchor video, its negative and positive pairs. We calculate this loss by iterating over all possible sets of such videos and their summaries. The parameter $\tau$ is the smoothing factor of the loss functions. It is a hyperparameter of our approach. In order to perform the training of such a pipeline successfully, we need to define the distance function $\text{dist}(\cdot, \cdot)$

\begin{figure*}[t!]
  \centering
    \includegraphics[width=\linewidth]{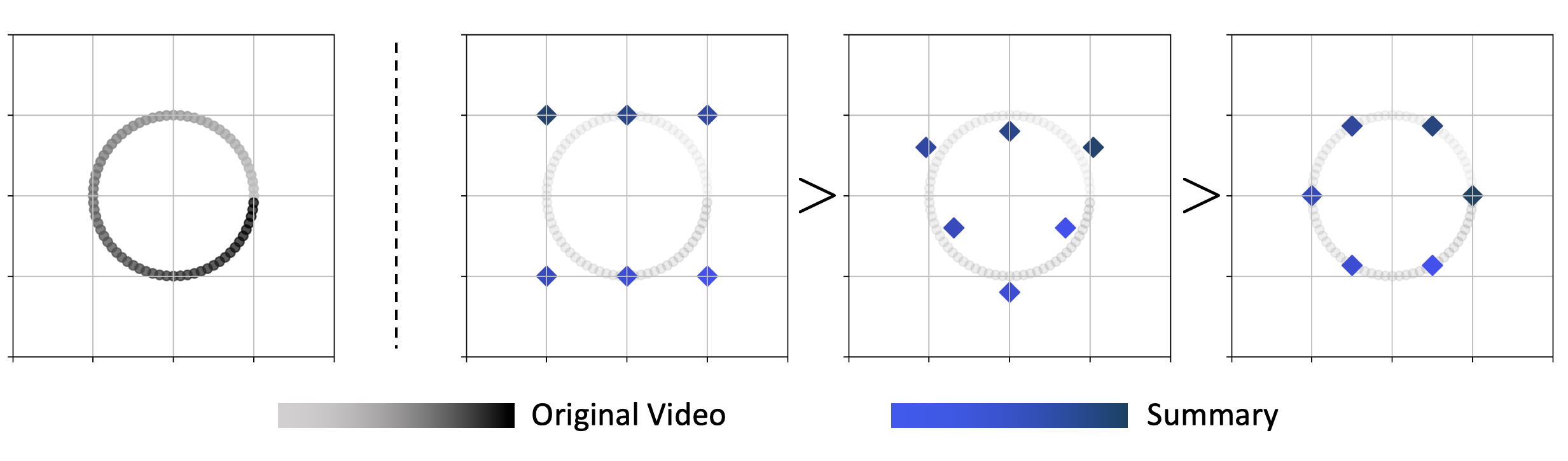}
  \caption{Left: a sample video as a set of short clips forms a circle in some hidden space. Right: Three sets of summaries that yield the same distance function when they are time-averaged. However, these summaries have different distances from the whole video if the distance function is given as in Equation \ref{eq:csum_three} which leads to a uniform distribution of clips.}
  \label{fig:csum_distance_function}
\end{figure*}

\subsection{Clip-Contrastive Distance Function}
Let us consider two sequences of $D$-dimensional vectors represented as matrices:
\begin{equation}
    \label{eq:csum_seuqences}
    \bm{X} = \{X_{ij} \}_{i,j=1}^{D,t} \quad \quad  \bm{Y} = \{Y_{ij} \}_{i,j=1}^{D,T}
\end{equation}
where $t$ and $T$ and the lengths of the sequences.
In our case, these matrices are clip features for the summary and the original video and thus we assume $t < T$. 
A common approach for calculating the distance between two sequences \cite{badamdorj2022contrastive} is to compare their features averaged it time: $\bar{\bm{x}} = \sum_j X_{ij} / t'$ and $\bar{\bm{y}} = \sum_k Y_{ik} / T$. It can be done for example by calculating the scalar product of these vectors:
\begin{equation}
\begin{split}
    \label{eq:csum_base_distance}
    \text{dist}(\bm{X}, \bm{Y}) 
    &= \Big< 
    \sum_j X_{ij} / t',
    \sum_k Y_{ik} / T
    \Big> \\
    &= \frac{1}{T \cdot t} \sum_{ijk} X_{ij} Y_{ik}
\end{split}
\end{equation}

The main drawback of such a distance function is that it compares the videos just by calculating the discrepancy between the average value of their clips.
Thus, it does not take into account the diversity of clips within the video.

We suggest the following procedure for calculating the distance between two videos. Let us consider a parameter $n$ which we call the length of a sub-video. We consider each of the videos as a distribution of all possible sub-videos of the lengths $n$. And then we calculate the mathematical expectation of the distance calculated between a sub-video from the first video and a sub-video from the second video as follows:
\begin{equation}
\begin{split}
    \label{eq:csum_three}
    \text{dist}_n(\bm{X}, \bm{Y}) 
    &= 
    \mathbb{E}_{\bm{x}' \sim q_n(\bm{X})}
    \mathbb{E}_{\bm{y}' \sim q_n(\bm{Y})}
    <\bm{x}', \bm{y}'> \\
    &= \mathbb{E}_{\bm{x}' \sim q_n(\bm{X})}
    \mathbb{E}_{\bm{y}' \sim q_n(\bm{Y})}
    \sum_{j=1}^{n\times D} x_j' y_j' 
\end{split}
\end{equation}
where $q(\bm{X})$ is distribution of all possible sub-videos from $\bm{X}$ which have $n$ clips inside. Such a distance function will degrade to Equation \ref{eq:csum_base_distance} if we consider $n=1$. For all other cases, it will take into account not only the difference between the mean values of the video embeddings but also their distributions.

\subsection{Differentiable Summary Selection}

In order to perform end-to-end training of the proposed pipeline, we must make all of the steps differentiable. 
The feature extractor, the score predictor and the projector are parametrized with neural networks and are thus differentiable. 
A more sophisticated part of the pipeline is the module, which selects clips with the highest scores, the top-$k$ frame selector.
Given a set of frames $\{x_1, x_2, \dots, x_N\}$ and a set of corresponding scores $\{s_1, s_2, \dots s_N, \}$ the top-$k$ selector outputs a set of $k$ frames which have the highest scores.

Ranking a set of frames according to their scores is equivalent to choosing the frame with the highest score, then removing it from the set and repeating the operation again and again. Choosing the maximum, or the top-1 frame can be formalized as follows:
\begin{equation}
    x_\text{max} = \sum_j x_j \vmathbb{1}[s_j > s_i,\; \forall i\neq j]
    \label{eq:csum_max_frame}
\end{equation}
where $\vmathbb{1}[\dots]$ is the indicator function. The value of $x_\text{max}$ changes with jumps from $x_1$ to $x_2$ and so on when the corresponding scores dominate the other scores. If we fix all the scores but just one, and then vary it from $-\infty$ to $\infty$, the value of $x_\text{max}$ will change just once and this change will be a jump (see Figure \ref{fig:csum_differentiation}). Thus, the gradient of $x_\text{max}$ with respect to the varying score will remain 0 everywhere except for the point of the jump, where the gradient is undefined. Therefore, using this gradient value for back-propagation is not possible. 

By following \cite{goyal2018continuous} we use a relaxation of this step function. Equation \ref{eq:csum_max_frame} can be approximated as follows
\begin{equation}
    \begin{split}
        x_\text{max} &\approx
        \sum_j x_j \frac{\exp(\alpha s_j)}{\sum_i \exp(\alpha s_i)} \\
        &= \sum_j x_j \cdot \text{SoftMax}(\alpha s)_j
    \end{split}
    \label{eq:csum_max_frame_approx}
\end{equation}
The parameter $\alpha$ can be interpreted as the inverse of the width of the transition region and if $\alpha \rightarrow \infty$, then $\text{SoftMax}(\alpha s) \rightarrow \vmathbb{1}[\dots]$. 

To rank the set of frames, we step-by-step select the maximum element by using Equation \ref{eq:csum_max_frame_approx} and then subtract from the maximum score a large number, so that the same frame will not be selected on the next step. In order to minimize the computational complexity of such an operation, we follow the approach proposed in \cite{pietruszka2021successive} and compare pairs of frames. Thus, the processing time growth logarithmically with the number of frames.

\begin{table*}[t]
    \begin{center}
    \begin{tabular}{@{}l|c|c|ccccccc@{}}
    \toprule
    \multirow{2}{*}{Model} & \multirow{2}{*}{Source} & \multirow{2}{*}{Supervised} & \multicolumn{3}{c}{TVSum} &                   & \multicolumn{3}{c}{SumMe}                     \\ 
                    &       &                             & $F_1$            & $\tau$           & $\rho$  &         & $F_1$             & $\tau$           & $\rho$            \\ \midrule
    vsLSTM & \cite{zhang2016video}       & \cmark                           & 54.2          & -             & -      &       & 37.6          & -             & -             \\
    dppLSTM & \cite{zhang2016video}      & \cmark                           & 54.7          & -             & -      &       & 38.6          & -             & -             \\
    VASNet & {\cite{fajtl2018summarizing}}          & \xmark                          & 61.4          & 0.16          & 0.17     &     & 49.7          & 0.16          & 0.17          \\
    MSVA & {\cite{ghauri2021MSVA}}        & \cmark                           & 62.8          & \textbf{0.19} & \textbf{0.21}& & 54.4          & 0.20          & \textbf{0.23} \\
    \midrule
    CSUM vsLSTM, &Ours     & \xmark                          & 59.0          & -             & -        &     & 41.0          & -             & -             \\ 
    CSUM dppLSTM & Ours     & \xmark                         & 60.5          & -             & -        &     & 44.2          & -             & -             \\ 
    CSUM VASNet & Ours      & \xmark                          & 62.7          & 0.17          & 0.17     &     & 52.1          & 0.16          & 0.17          \\ 
    CSUM MSVA & Ours        & \xmark                          & \textbf{63.9} & \textbf{0.19} & 0.20         & & \textbf{58.2} & \textbf{0.22} & \textbf{0.23} \\ 
    \bottomrule
    \end{tabular}
    \end{center}
    \caption{Experimental results on the TVSum and SumMe dataset. The reported metrics are F1-score, Spearman and Kendall correlation coefficients. We compare various backbone models with default training regimes and the same model trained with our contrastive approach. The best results are in \textbf{bold}.}
    \label{tab:csum_tvsum_summe}
\end{table*}

\vspace{10mm}

\section{Experiments}
\label{sec:csum_experiments}

\begin{figure*}[t]
    \centering
    \includegraphics[width=\linewidth]{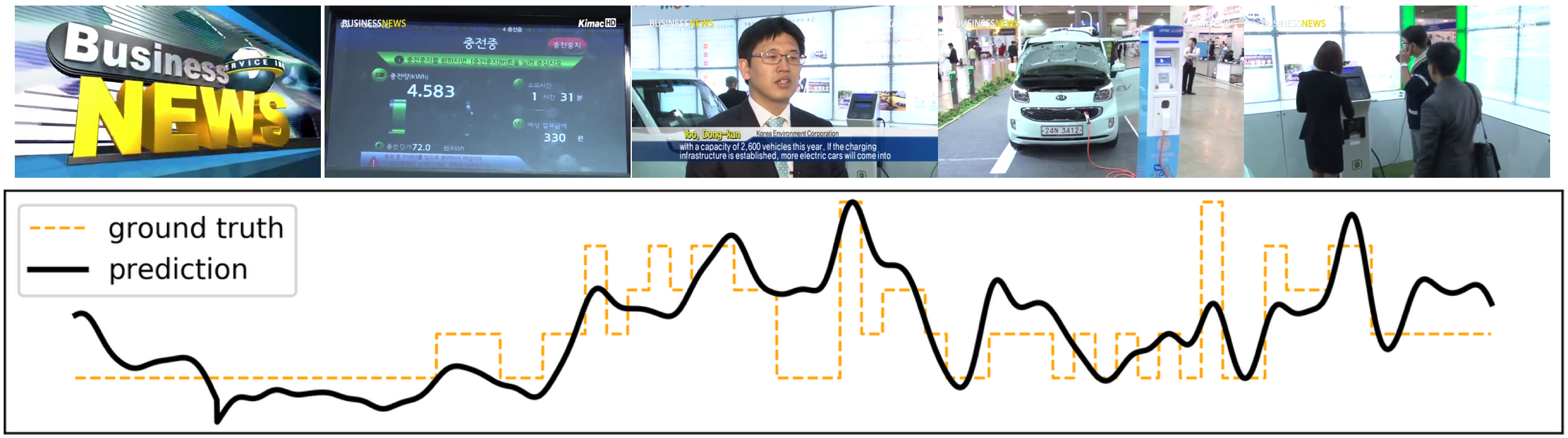}
    \caption{Top row: A visual example of the video summary extracted with our method. Bottom row: human-annotated ground truth importance scores and the importance scores trained with our contrastive learning method.}
    \label{fig:csum_qualitative}
\end{figure*}

\begin{table*}[t]
    \begin{center}
    \begin{tabular}{l|c|c|c|cccccc|c}
    \toprule
    Model       & Source          & Supervised & Audio  & Dog           & Gym.    & Park.       & Skat.       & Ski.        & Surf.       & Avg       \\ \midrule
    LSVM &\cite{sun2014ranking}    &  \cmark  & \xmark & 60.0          & 41.0          & 61.0          & 62.0          & 36.0          & 61.0          & 53.6          \\
    LIM-S &\cite{xiong2019less}    &  \cmark & \xmark & 57.9          & 41.7          & 67.0          & 57.8          & 48.6          & 65.1          & 56.4          \\
    SL-Module& \cite{xu2021cross} &  \cmark& \xmark & \textbf{70.8} & 53.2          & 77.2          & \textbf{72.5} & 66.1          & 76.2          & 69.3          \\
    CHD & \cite{badamdorj2022contrastive}        &  \xmark& \xmark & 60.6          & 71.1 & 74.2 & 49.8         & 68.2 & 68.5 & 65.4 \\
    CSUM UMT & Ours         &  \xmark& \xmark & 60.9         & 70.2 & 73.8 & 63.2          & 70.0 & 71.4 & 68.3\\
    CSUM UMT & Ours         &  \cmark& \xmark & 64.8          & \textbf{73.6} & \textbf{79.9} & 70.5          & \textbf{71.5} & \textbf{80.0} & \textbf{73.3} \\ \midrule
    MINI-Net &\cite{hong2020mini}  &  \cmark& \cmark & 58.2          & 61.7          & 70.2          & 72.2          & 58.7          & 65.1          & 64.4          \\
    TCG &\cite{ye2021temporal}      &  \cmark& \cmark & 55.4          & 62.7          & 70.9          & 69.1          & 60.1          & 59.8          & 63.0          \\
    Joint-VA & \cite{badamdorj2021joint} &  \cmark& \cmark & 64.5          & 71.9          & 80.8          & 62.0          & 73.2          & 78.3          & 71.8          \\
    UMT &\cite{liu2022umt}     &  \cmark& \cmark & 65.9          & \textbf{75.2} & \textbf{81.6} & 71.8          & 72.3          & 82.7          & 74.9          \\
    CSUM UMT & Ours         &  \cmark& \cmark & \textbf{66.1} & 75.1          & \textbf{81.6} & \textbf{71.9} & \textbf{73.0} & \textbf{82.8} & \textbf{75.1} \\
    \bottomrule
    \end{tabular}
    \end{center}
    \caption{Experimental results on the YouTube Highlights benchmark. The reported metric is mAP in percents. We compare both the methods which use the audio information from the video and the methods which rely on the visual features only. For both categories the best performing models are in \textbf{bold}.}
    \label{tab:csum_youtube_results}
\end{table*}

In this section, we evaluate the quality of video summarizations learned with the proposed method. We conduct experiments on several datasets and with several backbone models to demonstrate that the proposed method generalizes well for various video summarization setups. Next, we present qualitative examples of extracted video summarizations. Finally, we provide an ablation study on the hyper-parameters of our method.

\paragraph{Datasets} We conduct experiments with 3 datasets: TVSum \cite{song2015tvsum}, SumMe \cite{gygli2014creating} and YouTube Highlights \cite{sun2014ranking} datasets. The TVSum dataset consists of 50 videos from 10 categories from \cite{smeaton2006trecv}. In TVSum each video has frame-level importance scores annotated by 20 users. Importance scores range from 1 to 5, where 5 denotes the highest importance. The SumMe dataset includes 25 short videos of various events such as cooking or sports. Each video is attributed with frame-level importance scores. The YouTube Highlights dataset contains videos divided into 6 categories with around 1000 videos of various lengths available for each domain. For each video, there is a ground truth highlight in a form of a sequence of consecutive frames summarizing the content of the video in the best way.

\paragraph{Evaluation} To quantitatively evaluate the quality of extracted summaries we employ 5-fold cross-validation with an average F1-score across the splits. The cross-validation splits are the same as in \cite{ghauri2021MSVA}. The average F1-score over videos in the dataset is reported. As noted in \cite{otani2019rethinking} F1-score has certain limitations. We thus also adopt Spearman’s correlation ($\rho$) and Kendall correlation ($\tau$) coefficients between the summaries predicted by the models and ground truth summaries. For the YouTube Highlights dataset we perform a summary evaluation as a task of highlight detection in time. We thus employ mean average precision (mAP) as a known detection metric. The final mAP score is computed over [0.5:0.05:0.95] IoU thresholds.

\paragraph{Backbone models} To demonstrate that our method generalizes for various setups, we conduct experiments with several known backbone models: Video-LSTM and bi-directional Video-LSTM \cite{zhang2016video}, LSTM with attention \cite{fajtl2018summarizing}, Multi-Source Visual Attention model \cite{ghauri2021MSVA} and multi-modal Transformers \cite{liu2022umt}. For our experiments, we leave the backbone architecture unchanged and only modify the training pipeline of the models.

\subsection{Summarization performance}

We start with summarization experiments on the TVSum and SumMe datasets. Here we evaluate the proposed contrastive learning approach with various feature extraction backbone models. We use the proposed differentiable top-k summary extractor during training. During the inference stage, we simply select $N$ frames with the highest predicted scores. The results are reported in Table \ref{tab:csum_tvsum_summe}.

As can be seen from Table \ref{tab:csum_tvsum_summe}, using the proposed approach results in significant improvement for all of the baseline models. Notably, video LSTM (v-LSTM) enjoys a $4.8\%$ improvement in F1-score on the TVSum dataset, given that the proposed contrastive training does not use labels compared to its default supervised training regime. Also, our approach outperforms the previous best-performing unsupervised method VASNet \cite{fajtl2018summarizing} by $1.1\%$ on TVSum and by up to $2.4\%$ on SumMe, when the performance is measured with the F1-score. In terms of Spearman and Kendall correlation coefficients, our unsupervised method performs on par with the supervised models.

This experiment demonstrates that our contrastive learning approach generalizes well for various backbone architectures and for various datasets. Without using any labels, we are able either to match or to outperform existing supervised methods.

\subsection{Ablation studies}

In this section, we ablate the top-k selection algorithm and the window parameter $n$ in Equation \ref{eq:csum_three}. We also investigate if our top-k selector can robustly distribute importance scores regardless of the number of input frames.

For top-k differentiable selection ablation, we compare Sinkhorn \cite{xie2020differentiable}, Perturbed top-k \cite{cordonnier2021differentiable} and Successive Halving \cite{pietruszka2021successive} algorithms. As can be seen from Table \ref{tab:csum_ablation_top_k} the choice of top-k influences the final performance with Successive Halving delivering the best results for the MSVA backbone \cite{ghauri2021MSVA} on the TVSum dataset. We thus chose to use Successive Halving in all of the experiments.

\begin{figure}[h!]
    \centering
    \includegraphics[width=0.8\linewidth]{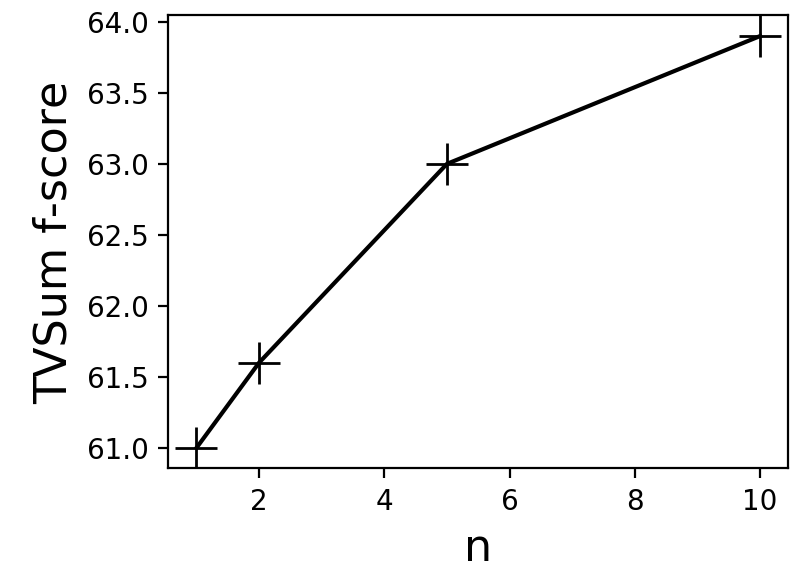}
    \caption{$F_1$-score of the MSVA backbone \cite{ghauri2021MSVA} on the TVSum dataset for various values of $n$ used in Equation \ref{eq:csum_three}.}
    \label{fig:csum_ablation_score}
\end{figure}

We next ablate the window parameter $n$ in Equation \ref{eq:csum_three}. Intuitively, $n$ is responsible for the granularity of the resulting video summarization, where lower values of $n$ result in higher granularity. As can be seen in Figure \ref{fig:csum_ablation_score}, the summarization quality benefits from higher values of $n$. That indicates that good video summaries should not be of the highest granularity. Thus, we use $n=10$ in all of the experiments.

\begin{figure}[t]
    \centering
    \includegraphics[width=0.8\linewidth]{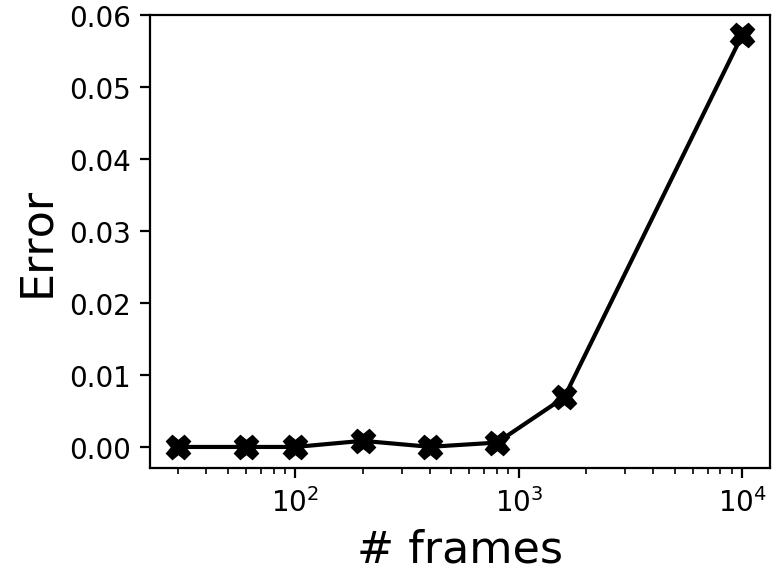}
    \caption{Normalized $L2$-error between the feature maps of top-10 frames selected with the proposed method and ground truth feature maps. On the x-axis is the total number of frames to extract the summary from.}
    \label{fig:csum_ablation_topk}
\end{figure}

Finally, we investigate if our differentiable top-k selector is robust with respect to the number of input frames. In Figure \ref{fig:csum_ablation_topk} we report how the normalized $L2$-error between the feature maps of top-10 frames selected with our method and ground truth feature maps depends on the number of input frames. The results suggest that even when the number of input frames is huge, the error does not exceed $0.06$. It indicates that the used differentiable top-k selector is robust with respect to the number of input frames.

\begin{table}[t]
    \begin{center}
    \begin{tabular}{lc}
    \toprule
    Top-$k$ method       & $F_1$-score     \\ \midrule
    Sinkhorn \cite{xie2020differentiable}        & $63.1 \pm 0.4$   \\
    Perturbed \cite{cordonnier2021differentiable}        & $62.9 \pm 0.5$ \\ 
    Successive Halving \cite{pietruszka2021successive} & $63.4 \pm 0.3$\\ \bottomrule
    \end{tabular}
    \end{center}
    \caption{$F_1$-score of the MSVA model \cite{ghauri2021MSVA} with different top-k selection mechanism on the TVSum dataset. A Successive Halving algorithm performs the best.}
    \label{tab:csum_ablation_top_k}
\end{table}

\subsection{Highlight detection}

We consider highlight detection as a special case of the summarization task, i.e. the highlight is a top-1 summary extraction coupled with surrounding context frames. Practically, to detect a highlight from a full-video, we prepossess summarization scores with Gaussian smoothing to enforce temporal continuity. After that, we extract a top-score frame with the surrounding frames with high enough scores to serve as one highlight. In the video summarization formulation, a clip is selected if it is among the fixed number of top-rated clips. For highlight detection, we select parts of the video which have score higher than a hyperparameter $\Theta$. We choose $\Theta$ by maximizing the mAP metric on a holdout set.

We conduct experiments in both supervised and unsupervised scenarios. For supervised highlight detection, we first pre-train the models with the proposed contrastive approach for 20 epochs and then fine-tune it for 50 epochs using the loss described in \cite{liu2022umt}. Evaluating fine-tuned representation is a standard procedure in contrastive learning \cite{chen2020simple,chen2021exploring}. For unsupervised highlight detection, we directly use the scores after 20 epochs of contrastive training.

We present the results for the cases when audio features are available and when they are not. In the supervised scenario, as can be seen from Table \ref{tab:csum_youtube_results}, our method (CSUM UMT) outperforms the competitive approaches or performs on par. In particular, for the no-audio case our method delivers more than $5\%$ improvement relative to the best-performing non-contrastive method \cite{xu2021cross}. With the audio information included, our method slightly outperforms the baseline UMT model, when the only modification being made is the contrastive pre-training used. In the unsupervised case, our method delivers more than $4\%$ increase in mAP score with respect to the previous best performing unsupervised method from \cite{badamdorj2022contrastive}. Also, the mAP score of our unsupervised model is only $1\%$ behind \cite{xu2021cross}, which fully relies on training with labels. 

We conclude that our contrastive approach is very competitive with existing methods, even when comparing our unsupervised with previous supervised results.

\subsection{Qualitative evaluation}
In Figure \ref{fig:csum_qualitative} we present an example of the video summarization of a sequence from the SumMe dataset trained with our contrastive framework and differentiable top-k. We can see that the predicted importance score can detect the regions of both low and high significance.

\section{Discussion}
\label{sec:csum_conclusion}

In this work, we propose a novel approach for unsupervised video summarization. We start by formulating the requirements for a good video summary: representatives, sparsity, and diversity. We observe that the contrastive learning framework naturally includes representatives and diversity. For sparsity, we propose a differentiable top-k selector based on predicted frame-level scores, where the importance is inherently distributed only among top-k input frames. This allows stepping away from comparing mean feature vectors, which may result in sub-optimal solution space, during the contrastive learning stage. Our approach does not rely on a specific kind of backbone; we experimentally show that it generalizes well for various architectures and summarization scenarios.

{\small
\bibliographystyle{ieee_fullname}
\bibliography{egbib}

\begin{thebibliography}{10}\itemsep=-1pt

\bibitem{badamdorj2021joint}
Taivanbat Badamdorj, Mrigank Rochan, Yang Wang, and Li Cheng.
\newblock Joint visual and audio learning for video highlight detection.
\newblock In {\em Proceedings of the IEEE/CVF International Conference on
  Computer Vision}, pages 8127--8137, 2021.

\bibitem{badamdorj2022contrastive}
Taivanbat Badamdorj, Mrigank Rochan, Yang Wang, and Li Cheng.
\newblock Contrastive learning for unsupervised video highlight detection.
\newblock In {\em Proceedings of the IEEE/CVF Conference on Computer Vision and
  Pattern Recognition}, pages 14042--14052, 2022.

\bibitem{caron2021emerging}
Mathilde Caron, Hugo Touvron, Ishan Misra, Herv{\'e} J{\'e}gou, Julien Mairal,
  Piotr Bojanowski, and Armand Joulin.
\newblock Emerging properties in self-supervised vision transformers.
\newblock In {\em Proceedings of the IEEE/CVF International Conference on
  Computer Vision}, pages 9650--9660, 2021.

\bibitem{chen2020simple}
Ting Chen, Simon Kornblith, Mohammad Norouzi, and Geoffrey Hinton.
\newblock A simple framework for contrastive learning of visual
  representations.
\newblock In {\em International conference on machine learning}, pages
  1597--1607. PMLR, 2020.

\bibitem{chen2021exploring}
Xinlei Chen and Kaiming He.
\newblock Exploring simple siamese representation learning.
\newblock In {\em Proceedings of the IEEE/CVF Conference on Computer Vision and
  Pattern Recognition}, pages 15750--15758, 2021.

\bibitem{cong2011towards}
Yang Cong, Junsong Yuan, and Jiebo Luo.
\newblock Towards scalable summarization of consumer videos via sparse
  dictionary selection.
\newblock {\em IEEE Transactions on Multimedia}, 14(1):66--75, 2011.

\bibitem{cordonnier2021differentiable}
Jean-Baptiste Cordonnier, Aravindh Mahendran, Alexey Dosovitskiy, Dirk
  Weissenborn, Jakob Uszkoreit, and Thomas Unterthiner.
\newblock Differentiable patch selection for image recognition.
\newblock In {\em Proceedings of the IEEE/CVF Conference on Computer Vision and
  Pattern Recognition}, pages 2351--2360, 2021.

\bibitem{dave2022tclr}
Ishan Dave, Rohit Gupta, Mamshad~Nayeem Rizve, and Mubarak Shah.
\newblock Tclr: Temporal contrastive learning for video representation.
\newblock {\em Computer Vision and Image Understanding}, 219:103406, 2022.

\bibitem{de2011vsumm}
Sandra Eliza~Fontes De~Avila, Ana Paula~Brandao Lopes, Antonio da Luz~Jr, and
  Arnaldo de Albuquerque~Ara{\'u}jo.
\newblock Vsumm: A mechanism designed to produce static video summaries and a
  novel evaluation method.
\newblock {\em Pattern recognition letters}, 32(1):56--68, 2011.

\bibitem{fajtl2018summarizing}
Jiri Fajtl, Hajar~Sadeghi Sokeh, Vasileios Argyriou, Dorothy Monekosso, and
  Paolo Remagnino.
\newblock Summarizing videos with attention.
\newblock In {\em Asian Conference on Computer Vision}, pages 39--54. Springer,
  2018.

\bibitem{fernando2017self}
Basura Fernando, Hakan Bilen, Efstratios Gavves, and Stephen Gould.
\newblock Self-supervised video representation learning with odd-one-out
  networks.
\newblock In {\em Proceedings of the IEEE conference on computer vision and
  pattern recognition}, pages 3636--3645, 2017.

\bibitem{gharbi2019key}
Hana Gharbi, Sahbi Bahroun, and Ezzeddine Zagrouba.
\newblock Key frame extraction for video summarization using local description
  and repeatability graph clustering.
\newblock {\em Signal, Image and Video Processing}, 13(3):507--515, 2019.

\bibitem{ghauri2021MSVA}
Junaid~Ahmed Ghauri, Sherzod Hakimov, and Ralph Ewerth.
\newblock Supervised video summarization via multiple feature sets with
  parallel attention.
\newblock 2021.

\bibitem{gong2014diverse}
Boqing Gong, Wei-Lun Chao, Kristen Grauman, and Fei Sha.
\newblock Diverse sequential subset selection for supervised video
  summarization.
\newblock {\em Advances in neural information processing systems}, 27, 2014.

\bibitem{gong2000video}
Yihong Gong and Xin Liu.
\newblock Video summarization using singular value decomposition.
\newblock In {\em Proceedings IEEE Conference on Computer Vision and Pattern
  Recognition. CVPR 2000 (Cat. No. PR00662)}, volume~2, pages 174--180. IEEE,
  2000.

\bibitem{goroshin2015unsupervised}
Ross Goroshin, Joan Bruna, Jonathan Tompson, David Eigen, and Yann LeCun.
\newblock Unsupervised learning of spatiotemporally coherent metrics.
\newblock In {\em Proceedings of the IEEE international conference on computer
  vision}, pages 4086--4093, 2015.

\bibitem{goyal2018continuous}
Kartik Goyal, Graham Neubig, Chris Dyer, and Taylor Berg-Kirkpatrick.
\newblock A continuous relaxation of beam search for end-to-end training of
  neural sequence models.
\newblock In {\em Proceedings of the AAAI Conference on Artificial
  Intelligence}, volume~32, 2018.

\bibitem{gygli2014creating}
Michael Gygli, Helmut Grabner, Hayko Riemenschneider, and Luc~Van Gool.
\newblock Creating summaries from user videos.
\newblock In {\em European conference on computer vision}, pages 505--520.
  Springer, 2014.

\bibitem{hadsell2006dimensionality}
Raia Hadsell, Sumit Chopra, and Yann LeCun.
\newblock Dimensionality reduction by learning an invariant mapping.
\newblock In {\em 2006 IEEE Computer Society Conference on Computer Vision and
  Pattern Recognition (CVPR'06)}, volume~2, pages 1735--1742. IEEE, 2006.

\bibitem{he2019unsupervised}
Xufeng He, Yang Hua, Tao Song, Zongpu Zhang, Zhengui Xue, Ruhui Ma, Neil
  Robertson, and Haibing Guan.
\newblock Unsupervised video summarization with attentive conditional
  generative adversarial networks.
\newblock In {\em Proceedings of the 27th ACM International Conference on
  multimedia}, pages 2296--2304, 2019.

\bibitem{hong2020mini}
Fa-Ting Hong, Xuanteng Huang, Wei-Hong Li, and Wei-Shi Zheng.
\newblock Mini-net: Multiple instance ranking network for video highlight
  detection.
\newblock In {\em European Conference on Computer Vision}, pages 345--360.
  Springer, 2020.

\bibitem{jing2020self}
Longlong Jing and Yingli Tian.
\newblock Self-supervised visual feature learning with deep neural networks: A
  survey.
\newblock {\em IEEE transactions on pattern analysis and machine intelligence},
  43(11):4037--4058, 2020.

\bibitem{kuanar2013video}
Sanjay~K Kuanar, Rameswar Panda, and Ananda~S Chowdhury.
\newblock Video key frame extraction through dynamic delaunay clustering with a
  structural constraint.
\newblock {\em Journal of Visual Communication and Image Representation},
  24(7):1212--1227, 2013.

\bibitem{li2018local}
Yandong Li, Liqiang Wang, Tianbao Yang, and Boqing Gong.
\newblock How local is the local diversity? reinforcing sequential
  determinantal point processes with dynamic ground sets for supervised video
  summarization.
\newblock In {\em Proceedings of the European Conference on Computer Vision
  (ECCV)}, pages 151--167, 2018.

\bibitem{liu2022umt}
Ye Liu, Siyuan Li, Yang Wu, Chang~Wen Chen, Ying Shan, and Xiaohu Qie.
\newblock Umt: Unified multi-modal transformers for joint video moment
  retrieval and highlight detection.
\newblock In {\em Proceedings of the IEEE/CVF Conference on Computer Vision and
  Pattern Recognition (CVPR)}, 2022.

\bibitem{mademlis2018regularized}
Ioannis Mademlis, Anastasios Tefas, and Ioannis Pitas.
\newblock Regularized svd-based video frame saliency for unsupervised activity
  video summarization.
\newblock In {\em 2018 IEEE International Conference on Acoustics, Speech and
  Signal Processing (ICASSP)}, pages 2691--2695. IEEE, 2018.

\bibitem{mahasseni2017unsupervised}
Behrooz Mahasseni, Michael Lam, and Sinisa Todorovic.
\newblock Unsupervised video summarization with adversarial lstm networks.
\newblock In {\em Proceedings of the IEEE conference on Computer Vision and
  Pattern Recognition}, pages 202--211, 2017.

\bibitem{mei20142}
Shaohui Mei, Genliang Guan, Zhiyong Wang, Mingyi He, Xian-Sheng Hua, and
  David~Dagan Feng.
\newblock L 2, 0 constrained sparse dictionary selection for video
  summarization.
\newblock In {\em 2014 IEEE international conference on multimedia and expo
  (ICME)}, pages 1--6. IEEE, 2014.

\bibitem{moskalev2022contrasting}
Artem Moskalev, Ivan Sosnovik, Fischer Volker, and Arnold Smeulders.
\newblock Contrasting quadratic assignments for set-based representation
  learning.
\newblock In {\em European Conference on Computer Vision}, 2022.

\bibitem{mundur2006keyframe}
Padmavathi Mundur, Yong Rao, and Yelena Yesha.
\newblock Keyframe-based video summarization using delaunay clustering.
\newblock {\em International Journal on Digital Libraries}, 6(2):219--232,
  2006.

\bibitem{ngo2003automatic}
Chong-Wah Ngo, Yu-Fei Ma, and Hong-Jiang Zhang.
\newblock Automatic video summarization by graph modeling.
\newblock In {\em Proceedings Ninth IEEE International Conference on Computer
  Vision}, pages 104--109. IEEE, 2003.

\bibitem{otani2019rethinking}
Mayu Otani, Yuta Nakashima, Esa Rahtu, and Janne Heikkila.
\newblock Rethinking the evaluation of video summaries.
\newblock In {\em Proceedings of the IEEE/CVF Conference on Computer Vision and
  Pattern Recognition}, pages 7596--7604, 2019.

\bibitem{petersen2022differentiable}
Felix Petersen, Hilde Kuehne, Christian Borgelt, and Oliver Deussen.
\newblock Differentiable top-k classification learning.
\newblock In {\em International Conference on Machine Learning}, pages
  17656--17668. PMLR, 2022.

\bibitem{pietruszka2021successive}
Micha{\l} Pietruszka, {\L}ukasz Borchmann, and Filip Grali{\'n}ski.
\newblock Successive halving top-k operator.
\newblock In {\em Proceedings of the AAAI Conference on Artificial
  Intelligence}, volume~35, pages 15869--15870, 2021.

\bibitem{rochan2018video}
Mrigank Rochan, Linwei Ye, and Yang Wang.
\newblock Video summarization using fully convolutional sequence networks.
\newblock In {\em Proceedings of the European conference on computer vision
  (ECCV)}, pages 347--363, 2018.

\bibitem{roy2021spatiotemporal}
Shuvendu Roy and Ali Etemad.
\newblock Spatiotemporal contrastive learning of facial expressions in videos.
\newblock In {\em 2021 9th International Conference on Affective Computing and
  Intelligent Interaction (ACII)}, pages 1--8. IEEE, 2021.

\bibitem{smeaton2006trecv}
Alan~F. Smeaton, Paul Over, and Wessel Kraaij.
\newblock Evaluation campaigns and trecvid.
\newblock In {\em Proceedings of the 8th ACM International Workshop on
  Multimedia Information Retrieval}, MIR '06, page 321–330. Association for
  Computing Machinery, 2006.

\bibitem{song2015tvsum}
Yale Song, Jordi Vallmitjana, Amanda Stent, and Alejandro Jaimes.
\newblock Tvsum: Summarizing web videos using titles.
\newblock In {\em 2015 IEEE Conference on Computer Vision and Pattern
  Recognition (CVPR)}, pages 5179--5187, 2015.

\bibitem{sun2014ranking}
Min Sun, Ali Farhadi, and Steve Seitz.
\newblock Ranking domain-specific highlights by analyzing edited videos.
\newblock In {\em ECCV}, pages 787--802. Springer International Publishing,
  2014.

\bibitem{tschannen2019mutual}
Michael Tschannen, Josip Djolonga, Paul~K Rubenstein, Sylvain Gelly, and Mario
  Lucic.
\newblock On mutual information maximization for representation learning.
\newblock {\em arXiv preprint arXiv:1907.13625}, 2019.

\bibitem{xie2020differentiable}
Yujia Xie, Hanjun Dai, Minshuo Chen, Bo Dai, Tuo Zhao, Hongyuan Zha, Wei Wei,
  and Tomas Pfister.
\newblock Differentiable top-k with optimal transport.
\newblock {\em Advances in Neural Information Processing Systems},
  33:20520--20531, 2020.

\bibitem{xiong2019less}
Bo Xiong, Yannis Kalantidis, Deepti Ghadiyaram, and Kristen Grauman.
\newblock Less is more: Learning highlight detection from video duration.
\newblock In {\em Proceedings of the IEEE/CVF conference on computer vision and
  pattern recognition}, pages 1258--1267, 2019.

\bibitem{xu2021cross}
Minghao Xu, Hang Wang, Bingbing Ni, Riheng Zhu, Zhenbang Sun, and Changhu Wang.
\newblock Cross-category video highlight detection via set-based learning.
\newblock In {\em Proceedings of the IEEE/CVF International Conference on
  Computer Vision}, pages 7970--7979, 2021.

\bibitem{ye2021temporal}
Qinghao Ye, Xiyue Shen, Yuan Gao, Zirui Wang, Qi Bi, Ping Li, and Guang Yang.
\newblock Temporal cue guided video highlight detection with low-rank
  audio-visual fusion.
\newblock In {\em Proceedings of the IEEE/CVF International Conference on
  Computer Vision}, pages 7950--7959, 2021.

\bibitem{zhang2016video}
Ke Zhang, Wei-Lun Chao, Fei Sha, and Kristen Grauman.
\newblock Video summarization with long short-term memory.
\newblock In {\em European conference on computer vision}, pages 766--782.
  Springer, 2016.

\bibitem{zhao2017hierarchical}
Bin Zhao, Xuelong Li, and Xiaoqiang Lu.
\newblock Hierarchical recurrent neural network for video summarization.
\newblock In {\em Proceedings of the 25th ACM international conference on
  Multimedia}, pages 863--871, 2017.

\bibitem{zhao2018hsa}
Bin Zhao, Xuelong Li, and Xiaoqiang Lu.
\newblock Hsa-rnn: Hierarchical structure-adaptive rnn for video summarization.
\newblock In {\em Proceedings of the IEEE conference on computer vision and
  pattern recognition}, pages 7405--7414, 2018.

\bibitem{zhou2018deep}
Kaiyang Zhou, Yu Qiao, and Tao Xiang.
\newblock Deep reinforcement learning for unsupervised video summarization with
  diversity-representativeness reward.
\newblock In {\em Proceedings of the AAAI Conference on Artificial
  Intelligence}, volume~32, 2018.

\end{thebibliography}
}

\end{document}